\documentclass[letterpaper]{article} 
\usepackage{aaai25}  
\usepackage{times}  
\usepackage{helvet}  
\usepackage{courier}  
\usepackage[hyphens]{url}  
\usepackage{graphicx} 
\usepackage{natbib}  
\usepackage{caption} 
\usepackage{algorithm}
\usepackage{algorithmic}

\usepackage{newfloat}
\usepackage{listings}
\DeclareCaptionStyle{ruled}{labelfont=normalfont,labelsep=colon,strut=off} 
\floatstyle{ruled}
\newfloat{listing}{tb}{lst}{}
\floatname{listing}{Listing}
\usepackage[utf8]{inputenc} 
\usepackage[T1]{fontenc}    
\usepackage{url}            
\usepackage{booktabs}       
\usepackage{amsfonts}       
\usepackage{nicefrac}       
\usepackage{microtype}      
\usepackage[rgb]{xcolor}         
\usepackage{footnote}
\usepackage{graphicx}
\usepackage{array}
\usepackage{multirow}
\usepackage{amsmath}

\title{Relation Also Knows: Rethinking the Recall and Editing of Factual Associations in Auto-Regressive Transformer Language Models}
\author{
    Xiyu Liu\textsuperscript{\rm 1,2}, Zhengxiao Liu\textsuperscript{\rm 1,2}\thanks{Zhengxiao Liu and Zheng Lin are corresponding authors.}, Naibin Gu\textsuperscript{\rm 1,2}, Zheng Lin\textsuperscript{\rm 1,2}$^{\rm *}$,
    Wanli Ma\textsuperscript{\rm 3}, Ji Xiang\textsuperscript{\rm 1}, Weiping Wang\textsuperscript{\rm 1}
}
\affiliations{
    \textsuperscript{\rm 1}Institute of Information Engineering, Chinese Academy of Sciences, Beijing, China\\
    \textsuperscript{\rm 2}School of Cyber Security, University of Chinese Academy of Sciences, Beijing, China\\
    \textsuperscript{\rm 3}University of Electronic Science and Technology of China, Chengdu, China\\
    \{liuxiyu, liuzhengxiao, gunaibin, linzheng, xiangji, wangweiping\}@iie.ac.cn, uestc\_wlma@163.com
}

\usepackage{bibentry}

\begin{document}

\maketitle

\begin{abstract}
The storage and recall of factual associations in auto-regressive transformer language models (LMs) have drawn a great deal of attention, inspiring knowledge editing by directly modifying the located model weights. Most editing works achieve knowledge editing under the guidance of existing interpretations of knowledge recall that mainly focus on subject knowledge. However, these interpretations are seriously flawed, neglecting relation information and leading to the \emph{over-generalizing} problem for editing. In this work, we discover a novel relation-focused perspective to interpret the knowledge recall of transformer LMs during inference and apply it on single knowledge editing to avoid over-generalizing. Experimental results on the dataset supplemented with a new R-Specificity criterion demonstrate that our editing approach significantly alleviates over-generalizing while remaining competitive on other criteria, breaking the domination of subject-focused editing for future research. The code is available at https://github.com/sunshower-liu/RETS.
\end{abstract}
\section{Introduction}
Language models are often regarded as knowledge bases, storing factual associations in parameters which can be simply recalled through prompting~\cite{petroni-etal-2019-language,lester-etal-2021-power,jiang-etal-2020-know,roberts-etal-2020-much,petroni2020how,heinzerling-inui-2021-language,wang-etal-2021-generative}. For instance, for the factual association shown in triplet <\emph{Marco Reus}, \emph{citizen-of}, \textbf{O}> with the subject \emph{Marco Reus} and the relation \emph{citizen-of}, the object \textbf{O} can be obtained from the next token prediction of GPT-like language models given the prompt "\emph{Marco Reus is a citizen of}". Recent works investigate where factual knowledge is stored and how the factual knowledge is extracted from auto-regressive transformer LMs, suggesting that the feedforward MLP sublayer performs as key-value memories which is the key component for the storing and recall of factual associations~\cite{geva-etal-2021-transformer,geva-etal-2022-transformer,geva-etal-2023-dissecting}. 
\begin{figure}[ht]
    \centering
    \includegraphics[width=0.38\textwidth]{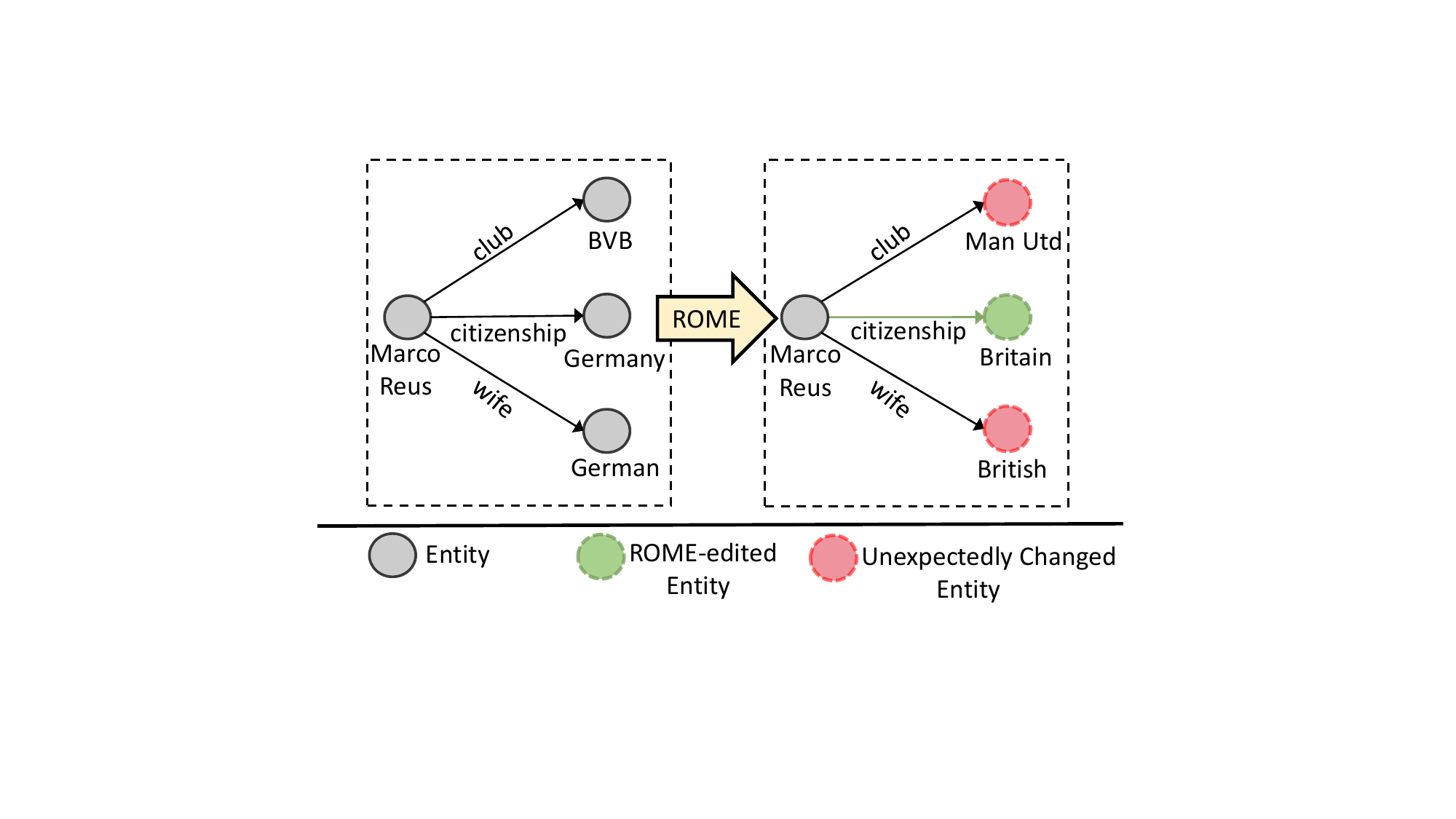}
    \caption{The over-generalizing problem. The circle in green denotes the correctly edited target entity and circles in red denote that the entities unrelated to target editing are also changed unexpectedly.}
    \label{fig1}
\end{figure}
The sight into the interpretation of auto-regressive transformer LMs makes renewing their knowledge by directly modifying the MLP weights possible, inspiring knowledge editing via the locate-then-edit paradigm that modifies the located weights~\cite{yao-etal-2023-editing,meng2022locating,Meng2022MassEditingMI,li2024pmet}. This sort of methods provide convenience for altering the behavior of models permanently on a small amount of facts while ensuring least side-effects, especially meaningful in the era of large language models.

\label{intro}

However, existing locate-then-edit methods suffer from various deficiencies~\cite{li2024unveiling,hoelscher-obermaier-etal-2023-detecting}. We note an over-generalizing problem that is serious in practical applications where unrelated relationships of the target editing subject experience unexpected alterations during the editing of a certain factual association. For example, the wife and other relationships of \emph{Marco Reus} predicted by the models will be changed to \emph{Britain}-related attributes while ROME~\cite{meng2022locating} edits the citizenship of \emph{Marco Reus} to \emph{Britain}, as illustrated in Figure~\ref{fig1}. This makes the contents generated by the edited models untrustworthy.

We conjecture that these locate-then-edit methods suffer from over-generalizing since they only focus on subjects and fail to take relations of factual associations into consideration during editing. Thus we firstly investigate what happens on relation tokens (e.g. "\emph{was born in}") in knowledge recall during inference to understand why previous works fail to take relations into account. The interpretation of knowledge recall involves \textbf{which positions} of tokens and \textbf{which layers} of auto-regressive transformer LMs primarily contribute to the prediction, and \textbf{what interpretable information} is encoded at the corresponding points. Through causal tracing~\cite{meng2022locating} for relations, we discover that the most contributing MLP and multi-head self attention MHSA sublayers for the propagation of relation representations appear at the last relation token. Furthermore, we analyze the trend of attributes rate and the target object ranking flow via the vocabulary lens~\cite{geva-etal-2021-transformer} of hidden representations at the identified last relation token across layers. Through the analysis results, we conclude with the relation-focused interpretation of knowledge recall that \textbf{relation-related attributes (i.e. relational knowledge)} are aggregated \textbf{from the first layer till middle-late layers at the last relation token} and that the target object token is extracted from the aggregated relational knowledge. We also validate the importance of MLP over MHSA for the aggregation of relational knowledge by the decline of attributes rate after blocking the MLP and MHSA sublayers respectively during inference. According to the investigation results, we notice that the inference of relations takes place at the last relation token and practically completes in middle-late layers. However, previous works achieve editing via modification of MLP in the middle-early layer at the last subject token, earlier than the inference of relations is completed and also unable to attend to the last relation token behind due to the nature of auto-regressive transformer LMs. As a result, previous locate-then-edit methods fail to take relational knowledge into account and tend to modify no matter what relationships of the target editing subject, leading to the over-generalizing problem.

In order to take relations into consideration, we propose to edit under the guidance of the novel relation-focused interpretation that modifies the MLP in the end of aggregation of relational knowledge (i.e. in the middle-late layer at the last relation token). Although simply editing at this point can attend to the subject, it loses the specificity on predictions of prompts with the same relation but different subjects with target editing (i.e. neighborhood subject prompts). Therefore, to make the hidden representations of neighborhood subject prompts more distinguishable at this point, we add an optimization target to the deduction of the weight modification to enhance the difference between such neighborhood prompts, constraining the editing to the certain subject. To sum up, we propose the \textbf{R}elation-focused \textbf{E}diting for auto-regressive \textbf{T}ransformer LMs with \textbf{S}ubject constraints (\textbf{RETS}) method to solve the problem of over-generalizing in single knowledge editing initially and verify the reliability of the novel relation-focused interpretation.

For evaluation, we supplement the \textsc{COUNTERFACT} ~\cite{meng2022locating} dataset with a new criterion Relation Specificity (i.e. R-Specificity) that measures the influence on unrelated facts of the edited subject. Experimental results on the supplemented dataset show that our editing method avoids over-generalizing by outperforming the state-of-the-art locate-then-edit methods over $30\%$ on Relation Specificity, while remaining competitive with the baselines on the previous criteria. Our strategy of single knowledge editing exhibits the most balanced performance overall and also validates the relation-focused interpretation on the recall of factual associations in auto-regressive transformer LMs, providing a novel perspective for future research on knowledge editing and the recall mechanism. 

\section{Related Work}
\label{related_work}
\paragraph{Interpretability of Transformer Language Models.}
We group the works that focus on the storage and recall mechanism of factual associations in GPT-like models~\cite{10.1145/3639372,luo2024understanding,kroeger2024large} based on concerning \emph{where} factual knowledge is stored and \emph{how} the knowledge is retrieved during inference.

The works concerning the storage of factual associations localize the knowledge captured by different transformer components\cite{Vaswani2017AttentionIA,kobayashi-etal-2020-attention,geva-etal-2022-transformer,geva-etal-2021-transformer}, suggesting that MLP sublayers, also known as the Feed-Forward Networks, act as key-value memories that store the factual associations~\cite{geva-etal-2021-transformer}. They further point out that each key-value pair of the MLP works as a sub-update that updates the token representation additively~\cite{geva-etal-2022-transformer}. Meanwhile, the multi-head self-attention MHSA layer is commonly known for its importance in linguistic capabilities~\cite{abnar-zuidema-2020-quantifying,katz2023visit,kobayashi2024analyzing}. These works provide a prerequisite for our preference to focus on MLP sublayers in knowledge recall.

The other works trace the information flow for the recall of factual associations during inference~\cite{meng2022locating,geva-etal-2023-dissecting,Hernandez2023LinearityOR}. One of them reveals the distinct set of middle-early MLP layers that significantly contribute to the factual predictions during processing the last-subject token via causal mediation analysis~\cite{meng2022locating}. Another work subsequently unveil that the representation at the last-subject position is enriched with subject-related attributes (i.e. subject knowledge) through middle-early MLP weights, but it ignores the existence of relational knowledge in knowledge recall~\cite{geva-etal-2023-dissecting}. Although some researchers~\cite{Hernandez2023LinearityOR} notice the role of relation, they explain the computation of a subset of relations as a well-approximated single-linear transformation on the subject representation, still limited to the subject-focused perspective that predicted tokens are extracted from subject knowledge and relations only function to map the subject knowledge to prediction.

As far as we know, none of the existing works about the interpretation of knowledge recall pays attention to the human-interpretable information of the relation representation, ignoring the relational knowledge. We are the first to explore the factual information recalled by the relation during inference.

\paragraph{Knowledge Editing.} Knowledge editing methods intend to alter the behavior of language models within the domain related to the edited fact, avoiding side-effects on unrelated facts~\cite{yao-etal-2023-editing,dai-etal-2022-knowledge,de-cao-etal-2021-editing,dong-etal-2022-calibrating,mitchell2022fast,hase2023does}. A line of locate-then-edit methods are proposed with the support of the recall mechanisms mentioned above, localizing a decisive MLP weight in middle-early layers at the last-subject position and directly modify it through rank-one model editing ROME~\cite{meng2022locating} for each single factual association. MEMIT~\cite{Meng2022MassEditingMI} improve ROME to be applicable on numerous edits simultaneously by spreading the update evenly over several middle MLP sublayers while processing the subject representation. PMET~\cite{li2024pmet} further obtains more precise FFN output at the last-subject position for editing by taking both MHSA and FFN information into consideration during optimization.

However, the state-of-the-art ROME-like methods primarily ignore the relation information while editing on the subject representation, exhibiting the deficiency of over-generalizing. Unlike these methods, we edit the auto-regressive transformer LMs on the relation representation while being able to take both the relation and the subject information into consideration. 

\section{Exploring the Role of Relation in Knowledge Recall}
\label{mechanism}
We firstly explore what happens on relations in knowledge recall through causal tracing and the analyses on vocabulary lens of hidden representations. 

\subsection{Background and Notation}
We give a description on the propagation within auto-regressive transformer LMs during inference.\footnote{The detailed description of the multi-head and nonessential layernorms and bias terms are omitted for simplicity.} Given an input text, these auto-regressive transformer LMs tokenize the input sequence into $t_1,t_2,...,t_N$ of length $N$ and embed them as vectors $h_1^0, h_2^0,...,h_N^0\in \mathbb{R}^d$ via the embedding matrix $E \in \mathbb{R}^{|\mathcal{V}|\times d}$ where the vocabulary size is $|\mathcal{V}|$. The models output the probability distribution of the next token $t_{N+1}\in\mathbb{R}^{|\mathcal{V}|}$ through transformer decoders of $L$ layers as follows:
\begin{equation}
    P(t_{N+1}|t_1,t_2,...,t_N)=\mathrm{softmax}(\phi(h_N^{L-1}+a_N^L+m_N^L))
\end{equation}
where $h_N^{L-1}$ is the residual hidden representation at $N$-th token from the layer ahead $L$-th layer, and $a_N^L$ and $m_N^L$ represent the outputs from $L$-th MHSA and MLP sublayers respectively. $\phi$ is the prediction head, mostly the multiplication as $\phi(x)=Ex$ or a trained linear layer. Generally, the hidden representation $h_i^l$, MLP output $m_i^l$ and MHSA output $a_i^l$ of layer $l\in{1,2,...,L}$ at token $t_i$ are calculated as follows:
\begin{equation}
    h_i^l = h_i^{l-1} + a_i^l + m_i^l
\end{equation}
\begin{equation}
\label{mhsa}
    a_i^l = (\sum_{j=1}^{N} \alpha_{i,j}^l\mathbf{v}^l(h_j^{l-1}))W_{O}^l
\end{equation}
\begin{equation}
\label{mlp}
    m_i^l = W_D^l\sigma (W_U^lI_{i}^l), \mathbf{v}(h_j^{l-1})=h_j^{l-1}W_V^l
\end{equation}

\begin{figure}[ht]
    \centering
    \includegraphics[scale=1.0]{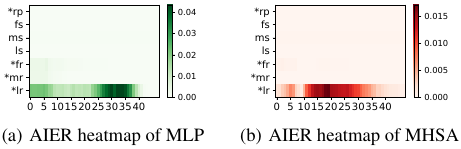}
    \centering
    \caption{Average Indirect Effect of Relation results for MLP and MHSA sublayers over 1000 facts on GPT2-XL. X-axis shows the layers. "rp" stands for the relation prefix in front of the subject (e.g. "The mother tongue language of" in the input prompt "The mother tongue language of Isabelle Breitman is"). "fs", "ms" and "ls" stand for the first-subject token, middle-subject tokens and last-subject token. "fr", "mr" and "lr" stand for the first-relation token, middle-relation tokens and last-relation token. "*" marks the intervened tokens in the corrupted run.}
    \label{fig_2}
\end{figure}

where $W_U^l\in\mathbb{R}^{d'\times d}$ and $W_D^l\in\mathbb{R}^{d\times d'}$ are the up-projection and down-projection weights of the MLP with the inner dimension of $d'$. $\sigma$ is the non-linear activation function. $I_{i}^l\in \mathbb{R}^{d}$ is the input vector of the MLP sublayer which is often assigned to $(h_i^{l-1} + a_i^l)$ for most auto-regressive transformer LMs but is assigned to $h_i^{l-1}$ for models with the parallel structure of MLP and MHSA. For the MHSA sublayer, $W_O^l\in \mathbb{R}^{d\times d}$ is the input weight matrix and the attention weight $ \alpha_{i,j}^l$ is given by:
\begin{equation}
    \alpha_{i,j}^l = \mathrm{softmax}(\frac{\mathbf{q}(h_j^{l-1})\mathbf{k}(h_j^{l-1})^T} {\sqrt{d}} + M_{ji}^l)
\end{equation}
\begin{equation}
    \mathbf{q}(h_j^{l-1})=h_j^{l-1}W_Q^l, ~\mathbf{k}(h_j^{l-1})=h_j^{l-1}W_K^l
\end{equation}
where $W_Q^l$, $W_K^l$, $W_V^l \in \mathbb{R}^{d\times d}$ are three projection matrices. $M_{ji}^l$ is the attention mask from $j$-th to $i$-th hidden representation in auto-regressive models. 

\subsection{Identifying Pivotal Positions of Relation}
We start by identifying which positions of relation tokens primarily contribute to knowledge recall. Here we display the results of GPT2-XL~\cite{radford2019language} with 48 layers (1.5B parameters). The results of GPT-J~\cite{mesh-transformer-jax} with 28 layers (6B parameters) and Llama-2~\cite{touvron2023llama2openfoundation} with 32 layers (7B parameters) are displayed in Appendix D, which both show similar trends with that of GPT2-XL.

\begin{figure*}[ht]
    \centering
    \includegraphics[scale=1.0]{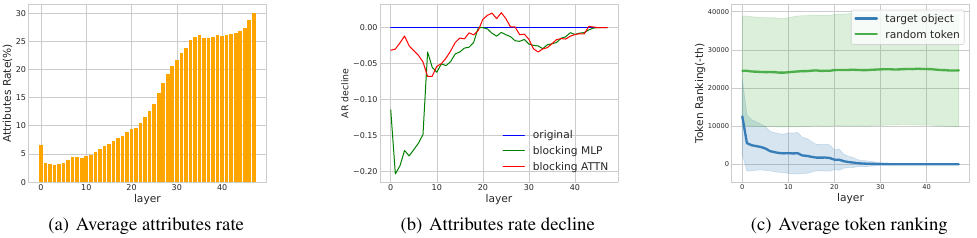}
    \centering
    \caption{The factual information detection on the vocabulary lens of the last-relation representation for GPT2-XL over 1000 prompts. (a) The average attributes rates as shown in yellow bars. (b) The average attributes rate decline at 48-th layer while blocking the MLP or MHSA sublayer respectively. (c) The average rankings of the target objects and random tokens.}
    \label{fig_3}
\end{figure*}

\paragraph{Method.} We utilize causal tracing~\cite{meng2022locating} to measure the importance of each inner activation for the relation tokens through three runs: a \emph{clean run}, a \emph{corrupted} run and a \emph{corrupted-with-restoration} run. In the clean run, a factual association prompt <$s,r$> is given to the model and the object $o$ is obtained from the output. All the clean internal activations (e.g. $m_i^l$ at token position $i$ in layer $l$) are cached during this run. Then, in the corrupted run, the embeddings of the relation $r$ is devastated by adding Gaussian noise $\mathcal{N}(0,\gamma)$ to them as $r'$ and the intervened input is sent to the model to obtain the probability for the original object $o$ as $\mathbb{P}(o|\mbox{<}s,r'\mbox{>})$. At last, in the corrupted-with-restoration run, the corrupted input $\mbox{<}s,r'\mbox{>}$ is still sent to the model but the cached clean hidden states are restored sequentially during inference, resulting in the probability $\mathbb{P}(o|\mbox{<}s,r'\mbox{>},x_i^l)$ for the output of resuming the activation $x_i^l$. The difference between our relation-focused causal tracing and previous subject-focused causal tracing lies in the corrupted run, where we add Gaussian noise to the relation tokens $r$ as $r'$ instead of the subject tokens. Thus, the contribution of each activation, namely Indirect Effect of Relation (\textbf{IER}), is calculated as follows:
\begin{equation}
    \mbox{\textbf{IER}} = \mathbb{P}(o|\mbox{<}s,r'\mbox{>},m_i^l) - \mathbb{P}(o|\mbox{<}s,r'\mbox{>})
\end{equation}

\paragraph{Results.} 
\begin{table}
 \setlength{\belowcaptionskip}{8pt}
 \renewcommand{\arraystretch}{1.2}
  \centering
  \begin{tabular}{cccc}
\hline \toprule[0.2mm]
\multirow{2}{*}{Position} & \multicolumn{3}{c}{Average AIER(\%)} \\ \cline{2-4} 
                                & GPT2-XL            & GPT-J    & Llama-2       \\ \hline
rp                              & 0.03               & 0.03     & 0.01           \\
fs                              & 0.04               & 0.04     & 0.01           \\
ms                              & 0.04               & 0.04     & 0.02           \\
ls                              & 0.04               & 0.04     & 0.01           \\
fr                              & 0.06               & 0.06     & 0.01            \\
mr                              & 0.06               & 0.06     & 0.02           \\
lr                              & \textbf{2.23}      & \textbf{3.15}    & \textbf{0.06}                \\ \bottomrule[0.2mm] \hline
\end{tabular}
\caption{Average AIER for different positions of tokens. The abbreviations here have the same meanings as in Figure~\ref{fig_2}.}
\label{AAIER}
\end{table}
Figure~\ref{fig_2} shows the average indirect effect of relation (AIER) heatmaps on MLP and MHSA sublayers for GPT2-XL. We note that for both MLP and MHSA, the most significant output representations are detected during inference at the last-relation token (also the last input token). Surprisingly, the AIER of relation prefixes ("rp" in Figure~\ref{fig_2}) contributes little to the prediction. All that matters for the entire relation information is the hidden representation at the last-relation token. Appendix A shows the AIER heatmaps on some specific cases. For quantitative comparisons, the average AIER across layers for tokens of certain positions is shown in Table~\ref{AAIER}. The average AIER at the last-relation position is far beyond that at any other position. This indicates that the MLP and MHSA sublayers are most active at the last-relation token which is the decisive position that process the potential factual information of the relation representation. Moreover, we can tell from Figure~\ref{fig_2}(a) that active MLP sublayers amass from the early layers to the middle-late site of the model, indicating that the update process for the relation representation is of long duration and finishes at the middle-late layers. Therefore, we conclude that the inference of relations primarily takes place at the last relation token from early layers to middle-late layers.


\subsection{Analyzing the Lens of Representations for Decisive Relation Positions}
\label{sec_lens}
After identifying the decisive position that dominate the inference of relation tokens, we explore what interpretable information is encoded in the hidden representations at last-relation position through the vocabulary lens~\cite{geva-etal-2021-transformer,geva-etal-2023-dissecting,luo2024understanding} on them.

\paragraph{Method.} For $h_i^l$ which indicates the hidden representation of $i$-th token at layer $l$, we map it to the distribution $\mathbf{p}_i^l$ over the vocabulary with the prediction head $\phi(h_i^l)$ which is the projection for prediction at the output layer. Thus, the hidden representation $h_i^l$ can be transformed into the ranking of tokens $r_i^l$ over the vocabulary as follows:
\begin{equation}
    \mathbf{p}_i^l = \mathrm{softmax}(\phi(h_i^l))
\end{equation}
\begin{equation}
    r_i^l = [v_1,v_2,...,v_{|\mathcal{V}|}] ~\mbox{where}~ \forall j>k, ~\mathbf{p}_i^l(v_j)> \mathbf{p}_i^l(v_k) 
\end{equation}
Here, $v_j$ stands for the token that ranked $j$-th in the whole vocabulary $\mathcal{V}$ according to $\mathbf{p}_i^l$. With the rankings, we analyze the hidden representations at last-relation position $N$ according to (i) the ranking of the predicted object $o$ given the input factual association <$s,r$> across layers (ii) the attributes rate metric~\cite{geva-etal-2023-dissecting} for the relation $r$ across layers, which measures the semantic relatedness between $r$ and the top ranked tokens $\mathbb{A}^l_{N}$ from the lens of $h_N^l$ at the last-relation token $t_N$. For automatic and convenient measurement, we collect a set $\mathbb{A}_{relation}$ containing 200 attributes for each relation $r$ via Wikidata Query Service\footnote{https://query.wikidata.org/} (see Appendix B for samples of the collection) and the attributes rate is computed as follows:

\begin{equation}
    \mathbf{AR}^l = \frac{|\mathbb{A}^l_{N} \cap \mathbb{A}_{relation}(r)|}{|\mathbb{A}^l_{N}|}
\end{equation}
where $\mathbf{AR}^l$ is the attributes rate of the relation representation at layer $l$ and $\mathbb{A}_{relation}(r)$ is the set of attributes related to $r$ of the input factual prompt. In practice, we select top $k=50$ tokens in each layer for $\mathbb{A}^l_{N}$ (see Appendix C for example).

\paragraph{Results.} 
Here we display the analysis results of GPT2-XL while the similar results of GPT-J can be found in Appendix E. Figure~\ref{fig_3}(a) presents the average attributes rates of the representation at last-relation position $h_{N}^l$ across layers. It shows that the average attributes rate has been rising significantly from layer $0$ (the first layer) till $36$-th layer and become stable afterwards. This trend indicates that the representation at last-relation position accumulates relation-related attributes from the early layers to the middle-late layers of the models, which is in accordance with the occurrence of the active MLP sublayers in Figure~\ref{fig_3}(a).

To further explore the importance of MLP and MHSA for the accumulation of relational knowledge respectively, we observe the average drops of attributes rate at 48-th layer while canceling the updates from MLP sublayers or MHSA sublayers at the last token respectively, results shown in Figure~\ref{fig_3}(b). It shows that blocking MLP leads to a much more significant drop in attributes rate than blocking MHSA across layers at the last token, indicating that MLP plays a much more important role in the enrichment of relational knowledge.
\begin{table}
 \renewcommand{\arraystretch}{1.2}
  \setlength{\belowcaptionskip}{8pt}
  \centering
    \begin{tabular}{ccc}
\hline \toprule[0.2mm]
Model   & Objects Included(\%) & $\rho$ \\ \hline
GPT2-XL & 68           & 0.97 \\
GPT-J   & 83           & 0.73 \\ \bottomrule[0.2mm] \hline
\end{tabular}
\caption{The percentage of facts where the objects are included in relation-related attributes and the Spearman rank coefficient $\rho\in[-1,1]$ between the average negative rankings of the objects and the average attributes rate.}
  \label{tab_AAIER}
\end{table}
Figure~\ref{fig_3}(c) plots the average rankings of the target objects and random tokens in the vocabulary distributions induced at the last-relation position. We can tell from the line charts that the average rankings of the target objects has been rising from early layers to middle-late layers, while that of random tokens remains low in all layers in comparison. This indicates the target objects are promoted to the final prediction gradually since the first layer of the models. Table~\ref{tab_AAIER} shows the proportion of the 1000 facts where the correctly predicted objects are included in corresponding $\mathbb{A}_{relation}(r)$ and the Spearman rank correlation coefficient between the average negative rankings of the objects and the average attributes rate of the representations of the last-relation position across layers. For GPT2-XL (GPT-J), $68\%$ ($83\%$) of correctly predicted objects are included in the corresponding $\mathbb{A}_{relation}(r)$ and the Spearman rank coefficient is $0.97$ ($0.73$), a strongly positive correlation between the extraction of the target objects and the accumulation of relation-related attributes. Thus, we conclude with the relation-focused interpretation that target objects are extracted from the relation-related attributes which are enriched at the last-relation token from early layers till middle-late layers and the MLP sublayers are essential in the update of relation representations. Under the guidance of this interpretation, we achieve editing by modifying the MLP sublayer in end of aggregation of relational knowledge (i.e. in the middle-late layer) with the relation representation (i.e. at the last-relation token) while taking subjects into account. 

\begin{figure}[ht]
    \centering
    \includegraphics[scale=0.31]{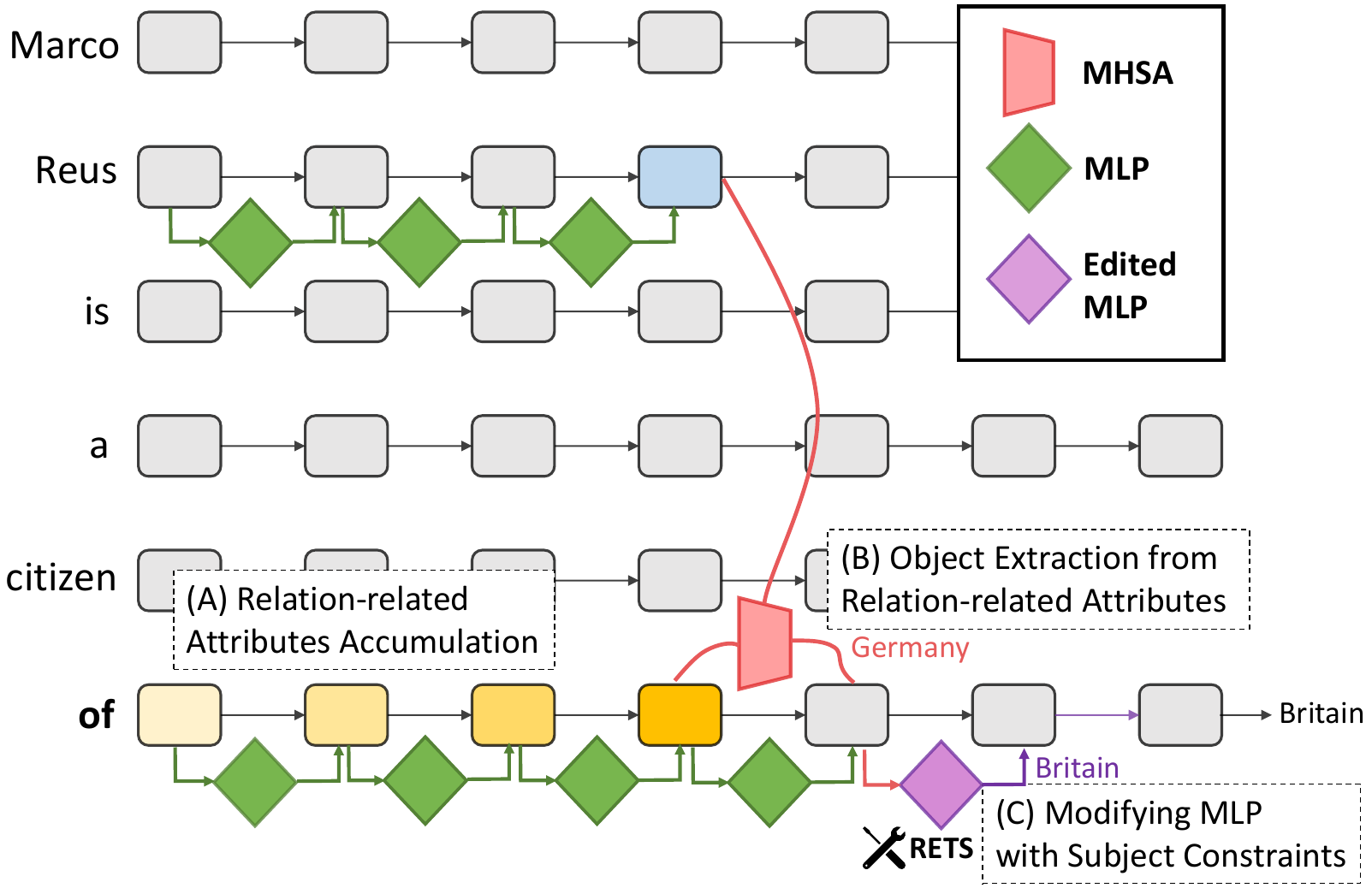}
    \caption{Our RETS method based on the relation-focused recall of factual associations. We reveal that the last-relation representation encodes relation-related attributes (A) which are accumulated until middle-late layers and (B) the predicted object is extracted from these attributes. Based on this relation-focused interpretation, we propose the RETS knowledge editing method that (C) modifies the middle-late MLP sublayer with the constraints of the subject.}
    \label{r-rome-fig}
\end{figure}

\section{Knowledge Editing from the Relation-focused Perspective}
\label{editing}
To further substantiate the importance of relational knowledge during inference, we apply the novel interpretation on knowledge editing to solve the over-generalizing problem. 

\subsection{Method: RETS}
We propose the Relation-focused Editing for auto-regressive Transformer models with Subject constraints (RETS) method that modifies the \textbf{middle-late} MLP sublayer with the hidden representation at the \textbf{last-relation position} while concerning the subject information, as illustrated in Figure~\ref{r-rome-fig}. The representation of the last-relation position is selected for its abundant factual information and the ability to attend to the subject tokens ahead. we choose the middle-late MLP sublayer for modification after accomplishing the attributes accumulation, constrained by information propagated from the subject tokens.

We give the formulization of the RETS method here. Requested to alter a factual association <$s,r,o$> to <$s,r,o^*$>, we choose to manipulate the forward pass at last-relation position $p_r$ by modifying the down-projection matrix $W_D^{l_e}$ of the MLP to $\tilde{W}_D^{l_e}$ in a middle-late layer $l_e$ which is in the end of the accumulation of relation-related attributes. The editing target is achieved by injecting $(k_*^{p_r},v_*^{p_r})$ into the associative memory and optimizing the objective function as follows:
\begin{equation}
    \tilde{W}^{l_e}k_*^{p_r}=v_*^{p_r}    
\end{equation}
\begin{equation}
    \mbox{minimize}~ ||\tilde{W}^{l_e}K-V||_F^2 + ||\tilde{W}^{l_e}K_{p_r}-V_{p_r}||_F^2
\end{equation}
\begin{equation}
\label{ori_mem}
    \mbox{minimize}~ ||W^{l_e}K-V||_F^2
\end{equation}

\begin{figure*}[htbp]
    \centering
    \includegraphics[scale=1.0]{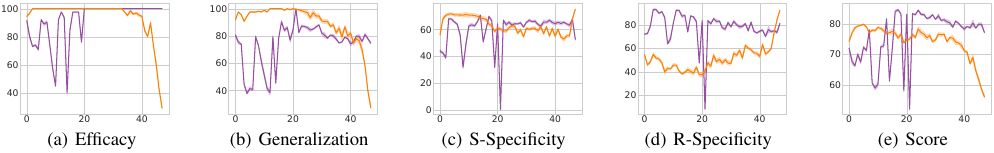}
    \centering
    \caption{The performance of RETS (purple lines) editing at the last-relation token on different layers (x-axis) compared with ROME (orange line) editing at the last-subject token for 50 prompts. Std deviation is shown in areas.}
    \label{layer_anal}
\end{figure*}

where $k_*^{p_r}$ is the average input hidden representation of $W_D^{l_e}$ with several prefixed prompts of <$s,r$> and $v_*^{p_r}$ is the output vector obtained by the optimization process (Appendix H). The first part of the objective function ensures the least change on the original key-value store $K=[~k_1~|~k_2~|~k_3~|~...~]$ and $V=[~v_1~|~v_2~|~v_3~|~...~]$. $W^{l_e}$ is the original associative memory  that memorizes the mapping from $K$ to $V$ by Eqn.~\ref{ori_mem}, solving as $W^{l_e}KK^T=VK^T$. The next part of the objective function is to emphasize the constraint for unrelated subjects. Since the subject representation propagates to the input vector $\sigma(W_U^lI^l_{p_r})$ of $W_D^{l_e}$ at $p_r$ (referring to Eqn.~\ref{mhsa} and \ref{mlp}), we collect a set of input vectors 
 $K_{p_r}=[~k^{p_r}_1~|~k^{p_r}_2~|~k^{p_r}_3|~...~]$ of several prompts with different subjects and their corresponding output vectors $V_{p_r}=[~v^{p_r}_1~|~v^{p_r}_2~|~v^{p_r}_3~|~...~]$ from $m_{p_r}^l$. Thus the facts of unrelated subjects are ensured the least alternation by minimizing $||\tilde{W}^{l_e}K_{p_r}-V_{p_r}||_F^2$, where we also have $W^{l_e}K_{p_r}K_{p_r}^T=V_{p_r}K_{p_r}^T$ for the original associative memory. By optimizing the objective function with all these constraints, we obtain the solution for $\tilde{W}^{l_e}$ as the new weight $W_D^{l_e}$ (see Appendix H for details):
 \begin{equation}
 \begin{split}
     \tilde{W}^{l_e} &= W^{l_e} + \Lambda^{p_r}((C+K_{p_r}K_{p_r}^T)^{-1}k_*^{p_s})^T \\
     &\approx W_D^{l_c} + \Lambda^{p_r}((C+K_{p_r}K_{p_r}^T)^{-1}k_*^{p_s})^T
\end{split}
 \end{equation}
where $C=KK^T$ is the constant estimated with the uncentered covariance of $k$ on a slice of Wikipedia corpus and $\Lambda^{p_r}$ is solved as $(v^{p_r}_*-W^{l_e}k^{p_r}_*)/((C+K_{p_r}K_{p_r}^T)^{-1}k_*^{p_r})^Tk_*^{p_r}$ which is proportional to the gap between the initial output vector $W^{l_e}k^{p_r}_*$ and the target output vector $v^{p_r}_*$. The linear memory $W^{l_e}$ is approximated by the model weight $W_D^{l_e}$.

\subsection{Experiments}
\label{main_exp}
\paragraph{Baselines and DataSets.}
Our editing experiments are mainly conducted on GPT2-XL and GPT-J for each single factual association. We also evaluate the basic editing performance on Llama-2 (7B). To compare with, we choose the methods for editing each single factual association as our baselines, including Constrained Fine-Tuning (FT+L)~\cite{zhu2020modifying}, the meta-learning method (MEND)~\cite{mitchell2022fast} which learns the update of model weights with additional networks, ROME~\cite{meng2022locating} on the subject representation in the early site and the improved precise model editing (PMET)~\cite{li2024pmet} which optimizes the parallel MHSA and MLP representations simultaneously.

For evaluation, we conduct experiments on $10,000$ ($2,000$) samples of \textsc{COUNTERFACT}~\cite{meng2022locating} dataset for GPT2-XL (GPT-J), $1,000$ samples of \textsc{COUNTERFACT} for Llama-2 and $10,000$ samples of Zero-shot Relation Extraction (zsRE)~\cite{levy-etal-2017-zero,meng2022locating} for GPT2-XL. We edit on the $36$-th layer of GPT2-XL, $18$-th layer of GPT-J and $23$-th layer of Llama-2. For \textsc{COUNTERFACT}, we supplement this dataset with unrelated facts of the same subject for each target editing and the corresponding metric Relation Specificity (\textbf{R-Specificity}) to measure the over-generalizing problem. To be specific, given the edited fact <$s_1,r_1,o_1^*$> and the unrelated fact with the same subject <$s_1,r_2,o_2$>, we test $\mathbb{P}[o_2] > \mathbb{P}[o_1^*]$ as the R-Specificity Score with the input prompt of <$s_1, r_2$>, computed similarly to the existing metrics. We also test the original metrics including the basic \textbf{Efficacy} accuracy Scores to measure the success rate of target editing, the \textbf{Generalization} accuracy score for generalization on the paraphrased statements, the renamed Subject Specificity \textbf{S-Specificity} accuracy score for specificity within neighborhood subjects. The advanced \textbf{Fluency} and \textbf{Consistency} scores measure the quality of generated full texts. Higher scores indicate better performance for all metrics. For zsRE, the metics and evaluation results are displayed in Appendix G. Details for the construction of R-Specificity samples and detailed settings are presented in Appendix F.
\begin{table*}[htbp]
\centering
\renewcommand{\arraystretch}{1.1}
\setlength{\tabcolsep}{1mm}{
\begin{tabular}{cccccc|cc}
\hline \toprule[0.2mm]
Editor  & Score & Efficacy & Generalization & S-Specificity & R-Specificity & Fluency & Consistency \\ \hline
GPT2-XL & 55.9      & 21.0     & 24.1           & 78.6          & 100.0             & 626.8   & 34.7        \\ \hline
FT-L    & 73.1      & 99.2     & \underline{47.8}           & 70.6          & 74.9          & 623.3   & 37.6        \\
MEND    & 63.2      & \underline{62.3}     & \underline{53.1}           & \underline{51.7}          & 85.6          & 603.7   & 32.7       \\
ROME    & 78.4      & 100.0    & 96.4          & 76.0          & \underline{41.1}          & 622.6   & 42.0      \\
PMET*    & 79.3      & 99.2     & 94.3          & 76.0          & \underline{47.6}          & 622.7   & 41.8       \\ \hline
\textbf{RETS}  & 79.7      & 100.0    & 71.5           & 68.6          & 78.5          & 577.4   & 32.6\\
\multicolumn{1}{r}{- w/o SC}  & 71.1      & 100.0    & 67.2           & 35.1          & 86.9          & 626.1   & 34.9\\ \hline \hline
GPT-J   & 53.2      & 13.7     & 15.3           & 83.7          & 100.0             & 621.7   & 29.7        \\ \hline
FT-L    & 79.3      &  99.6   &  \underline{47.4}           &  80.1          &  89.1         & 622.5   &  35.3   \\
MEND    & 75.3      & 96.8     & \underline{51.2}           & \underline{53.8}          & 99.2          & 620.4   & 32.2        \\
ROME    & 81.7      & 99.9    & 99.0           & 79.4          & \underline{48.5}          & 620.5   & 42.7        \\
PMET*    & 83.5      & 99.9     & 98.7           & 79.6          & \underline{55.6}          & 620.9   & 43.0        \\     \hline
\textbf{RETS}  & 80.7      & 100.0    & 74.2           & 65.5          & 83.3          & 542.4   & 29.2        \\ 
\multicolumn{1}{r}{- w/o SC}  & 74.1      & 100.0    & 82.0           & 23.7          & 90.7          & 618.1   & 34.9 \\ \hline \hline 
Llama-2 & 52.8  & 13.8     & 16.1           & 81.2          & 100.0 & - & -   \\ \hline
FT-L    & 55.6  & \underline{24.2}     & \underline{17.0}           & 81.6          & 99.7    & - & -      \\
ROME    & 81.1  & 99.9     & 93.4           & 77.4          & \underline{53.6}    & - & -      \\
\textbf{RETS}    & 82.1  & 98.3     & 74.6           & 72.3          & 83.1  & - & -        \\

\bottomrule[0.2mm] \hline
\end{tabular}}
\caption{The evaluation results on \textsc{COUNERTFACT} for GPT2-XL and GPT-J. The significantly failed values for the editing methods on basic criteria are underlined. "Score" shows the average value on the basic criteria: Efficacy, Generalization, S-Specificity and R-Specificity. "SC" stands for the subject constraints on our relation-focused editing. R-Specificity values for raw models are $100.0\%$ since the criterion is constructed according to the top token predictions of the raw models. *PMET is adjusted to accommodate to edit a single layer.}
\label{count_re}
\end{table*}

\paragraph{Evaluation Results.}
Table~\ref{count_re} shows the evaluation results on \textsc{COUNTERFACT}. We observe that the existing mainstream editing methods exhibit at least one deficiency. Even though existing ROME-like methods (ROME and PMET) perform well on most criteria, they experience obvious failure on the Relation Specificity. Our RETS method outperforms the ROME-like methods over $30\%$ on R-Specificity for both GPT2-XL and GPT-J while remaining competitive on other criteria, indicating that the relation-focused editing solves the over-generalizing problem for ROME initially. The evaluation result on recent Llama-2 also shows the same trend. The results on zsRE in Appendix G also demonstrate the competitiveness of RETS. An anecdotal case of RETS behaving correctly while ROME behaving erroneously on GPT2-XL is shown in Figure~\ref{case_study}.

The ablation of the subject constraints for editing COUNTERFACT leads to the $35\%$ decrease on Entity Specificity for GPT2-XL, which indicates the effectiveness of the subject constraints for the relation-focused editing. Despite of the results on Generalization and Entity Specificity where RETS loses about $20\%$ and $10\%$ respectively compared with the subject-centered editing methods, RETS exhibits the most balanced performance with its simple way of combining the subject information into editing, which shows the potential of editing from relation-focused perspective. The trade-off of performance is decided by the editing position (the last-relation token or the last-subject token) as expected. Editing at the last-relation position ensures minimal impact on unrelated relations (i.e. high R-Specificity) but loses much subject information (i.e. low S-Specificity), while the opposite is also true for editing at the last-subject position. The superiority of the relation-focused approach is that the decline of S-Specificity can be constrained by subject constraints whereas the subject-focused approach can hardly attend to
\begin{table}[ht]
    \centering
    \includegraphics[scale=0.31]{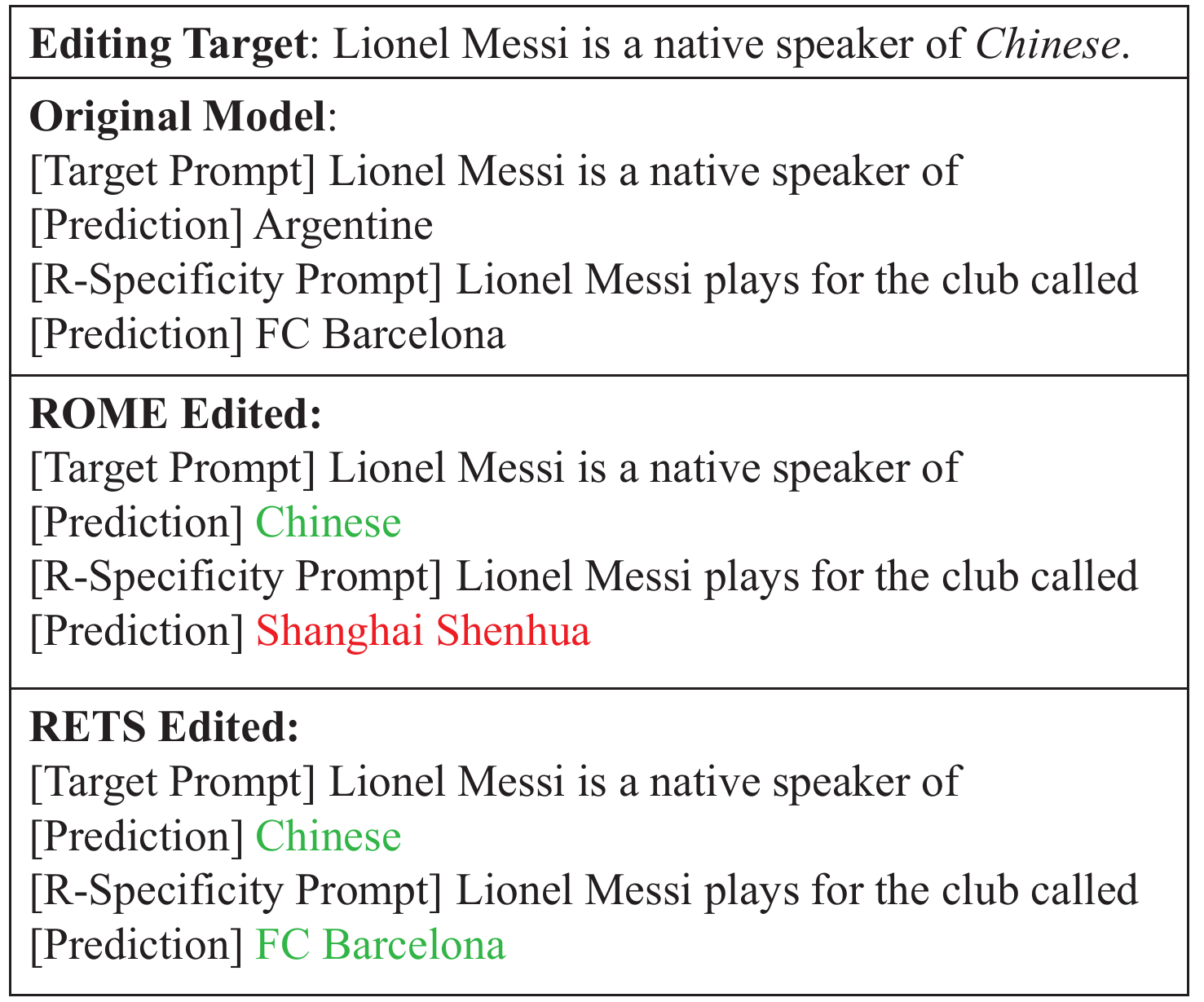}
    \caption{An anecdotal example of the correct behavior for RETS and the incorrect behavior for ROME on GPT2-XL. Predictions in red denote unexpectedly changed answers.}
    \label{case_study}
\end{table}
the relations. Detailed discussions can be referred to Appendix J.

\paragraph{Layer Analysis.}
We test the effectiveness of RETS while editing on different layers and compare it with the behavior of ROME which edits at the last-subject token . Figure~\ref{layer_anal} plots the performance on four criteria (a,b,c,d) and the average scores (e) across layers. The performance of RETS vibrates before middle layers and become stable after middle-late layers, validating our interpretation of relation-focused knowledge recall that the object is attracted from relation-related attributes which are accumulated before middle-late layers. RETS editing on middle-late layers shows more balanced performance than editing on any layer at the last-subject token where relation information behind hardly propagates to.

\section{Conclusion}
We discover the over-generalizing problem for previous subject-focused knowledge editing methods, and we solve this problem by further exploring the role of relations in knowledge recall. As a result, we unveil the factual information encoded for relations in auto-regressive transformer language models, and we propose the RETS single knowledge editing method based on the relation-focused interpretation. Our experiments demonstrate the effectiveness of RETS on solving the over-generalizing problem and provide the novel relation-focused perspective for future research on both the interpretation and editing of the auto-regressive transformer language models, breaking the domination of the subject-focused perspective.  

%
\section{Ethical Statement}
The goal of our work is to investigate and renew the outdated or mistaken knowledge decoded in transformer language models. However, we recognize inherent risks associated with potential malicious applications like injecting harmful information. Therefore, we emphasize the importance that language models be sourced exclusively from reputable and trustworthy providers and carefully use the contents generated by these models.

\section{Acknowledgements}
This work is supported by the National Natural Science Foundation of China (No. 62472419, No. 62472420).

\bibliography{rets}

\appendix
\section{Appendix}
\subsection{A. Cases for Causal Tracing on Relation Tokens}
\label{case_causal}
In Figure 7, we display the heatmaps of the average indirect effect of GPT2-XL and GPT-J on specific cases. The heatmaps show the centers of interval of 10 patched layers. All of the results exhibit the decisive position of the last relation token where the MLP outputs primarily contribute to the prediction and the most active MLP sublayers amass in either early or middle-late layers.
\begin{figure}[ht]
    \centering
    \includegraphics[scale=0.6]{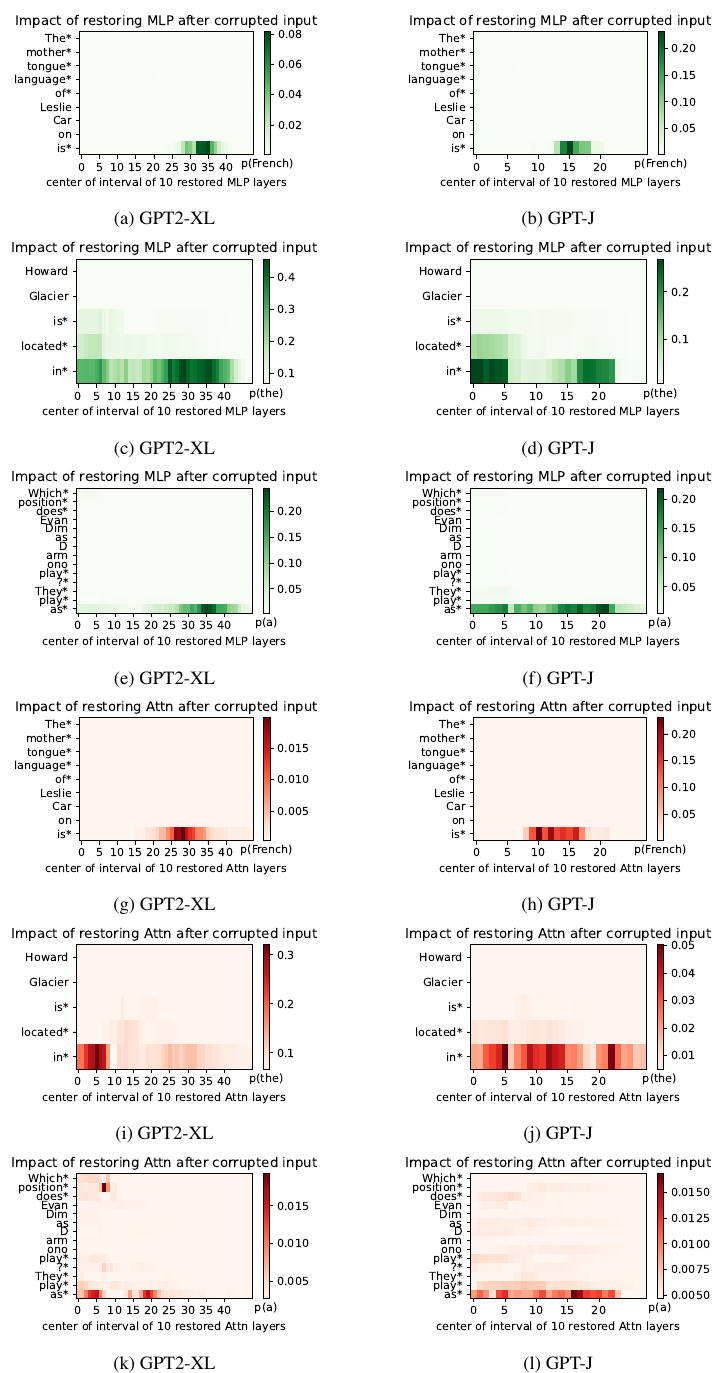}
    \centering
    \caption{The cases of Causal Tracing results of relation tokens on GPT2-XL and GPT-J.}
\end{figure}

\subsection{B. Samples of Collections for Ground-Truth Attributes}
\label{sample_attributes}
We collect the possible attributes of each relation for measuring the attributes rate automatically. The attributes are searched through SPARQL queries on Wikidata Query Service for the corresponding relation.  Here we display some samples of the collection in Table 4.
\begin{table*}[ht]
\centering
\renewcommand{\arraystretch}{1.2}
\caption{Samples of the ground-truth attributes.}
\begin{tabular}{p{4cm}p{2cm}p{7cm}}
\hline
Relation Prompt & Relation ID & Attributes \\ \hline
is a native speaker of    &   P103          &  French, Esperanto, German, Turkish, Icelandic, Portuguese, Japanese, Croatian, Catalan, Dutch, Russian, Urdu,
        English, ...       \\
is affiliated with the religion         &    P140         &   Islam, Buddhism, jansenism, Ganesha, Catholicism, cathedral, Christianity, anarchism, atheism, hedonism, communism, feminism, ...         \\
is a product of         &  P178           &     Google, Boeing, Airbus, Tim Berners-Lee, Wikimedia Foundation, Jimmy Wales, Larry Sanger, Intel, Volkswagen, ...  \\ \hline
\end{tabular}
\end{table*}

\subsection{C. Samples of Top Ranked Tokens through Vocabulary Lens}
\label{sample_toptok}
Each internal hidden representation in auto-regressive transformer language models can be viewed as a distribution over the vocabulary. Thus we investigate the top ranked tokens from the distribution through the vocabulary lens. Here in Table 5 we give some examples of the top ranked tokens induced by the hidden representation $h^l_N$ of layer $l$ at the last-relation token $t_N$. The top ranked tokens in early layers are nonsense while the top ranked tokens in middle-late layers are closely related to the the relation.
\begin{table*}[ht]
\centering
\renewcommand{\arraystretch}{1.2}
\caption{Samples of top ranked tokens induced by the hidden representation in GPT2-XL.}
\begin{tabular}{p{4cm}p{2cm}p{7cm}}
\hline
Prompt & Hidden State & Top-$10$ Ranked Tokens \\ \hline
Evan Dimas Darmono is a native speaker of    &   $h_N^{15}$          &  ' English', ' native', ' color', ' colour', ' language', ' flu', ' Am', ' N', ' am', ' Native'       \\
    & $h_N^{35}$ &  ' English', ' languages', ' Arabic', ' fluent', ' Spanish', ' Sanskrit', ' english', ' Hindi', ' Italian', ' Languages'        \\
Hersekzade Ahmed Pasha is affiliated with the religion         &    $h_N^{13}$         &   ' department', ' centre', 'ologist', ' establishment', ' center', "'s", ' novice', ' group', ' organization', ' elite'         \\
        &  $h_N^{37}$           &     ' sect', ' fundamentalist', ' Islam', ' Islamic', ' teachings', ' Ahmad', ' religion', ' extremist', ' group', ' of'  \\ \hline
\end{tabular}
\end{table*}

\subsection{D. Causal Tracing Results on GPT-J and Llama-2}
\begin{figure}[ht]
    \centering
    \includegraphics[scale=0.6]{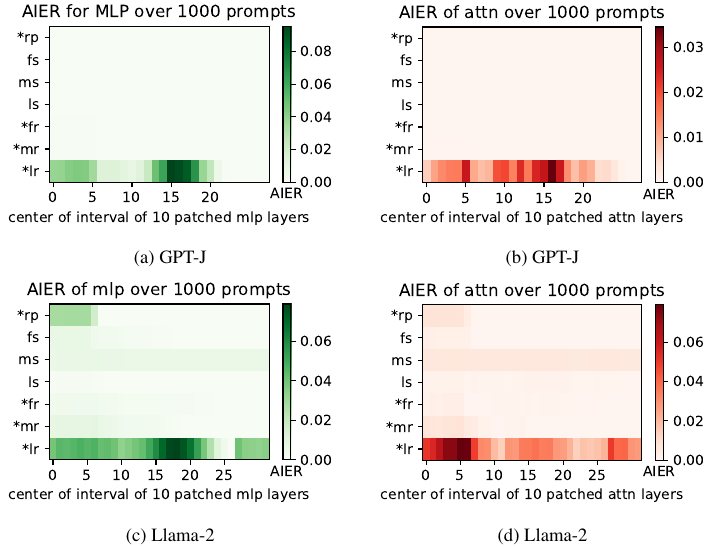}
    \centering
    \caption{The average causal tracing results of GPT-J and Llama-2.}
    \label{causal_j_llama}
\end{figure}

Here we display the average causal tracing results of GPT-J (6B) and Llama-2 (7B) in Figure~\ref{causal_j_llama}, both showing the similar phenomenon with GPT2-XL.  

\subsection{E. The Vocabulary Lens on GPT-J}
\begin{figure}[ht]
    \centering
    \includegraphics[scale=0.6]{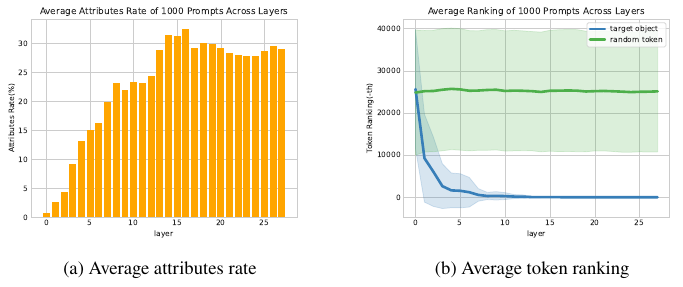}
    \centering
    \caption{The factual information detection on the vocabulary lens of the last-relation representation for GPT-J. (a) The average attributes rate of the representations.(b) The average rankings of the target objects and random tokens.}
    \label{appendix_lens}
\end{figure}

We display the average attributes rates and the average rankings of the tokens in GPT-J in Figure~\ref{appendix_lens}, which are consistent with the trends shown in GPT2-XL.

\subsection{F. Experimental Details}
\label{appendix_exp_detail}
\subsubsection{Metrics}
The existing metrics we use in our experiments are in accordance with Meng et al. 2022a. The original basic metrics include the basic \textbf{Efficacy} accuracy Scores to measure the success rate of target editing with $\mathbb{P}[o_1^*] > \mathbb{P}[o_1]$ where $o_1$ is the original object of <$s_1,r_1$>, the \textbf{Generalization} accuracy score for generalization on the paraphrased statement <$s_1, r_1'$> with $\mathbb{P}[o_1^*] > \mathbb{P}[o_1]$, the renamed Subject Specificity \textbf{S-Specificity} accuracy score for specificity within neighborhood subjects <$s_3, r_1$> with the original object $o_3$ that $\mathbb{P}[o_1^*] < \mathbb{P}[o_3]$.

We introduce the construction of the new criterion R-Specificity here. Firstly, we need to supplement the factual prompts that share the same subject but differ in the relation for each edit. To achieve this, we categorize all the subjects of the target factual associations in \textsc{COUNTERFACT} into \emph{person}, \emph{location}, \emph{organization} and \emph{product}. Then, for each category, we manually creates several prompts with the subject in the category and the relations that never appear in \textsc{COUNTERFACT}. With the supplemented prompts, we can measure how the prediction of the raw model for unrelated factual associations is distorted by editing. 

\subsubsection{Settings}
All the parameter settings of the baselines are consistent with Meng et al. 2022a and Meng et al. 2022b. To highlight our relation-centered perspective, the parameter settings of learning rate and other optimization parameters of RETS are also consistent with Meng et al. 2022a, except that we edit in $36$-th ($18$)-th layer for GPT2-XL (GPT-J). Evaluation experiments are conducted only once due to computational resource constraints.

For the implementaion of PMET for single editing, we simply adopt the calculation of $v$ vectors (Eqn. 4) from PMET and edit at a single layer instead of several layers as in Li et al. 2024a.

All of our experiments can be done on one NVIDIA A800 80GB GPU.

We use subject constraints to restrict the editing to the certain subject by adding a least-square loss on the objective function, where the loss is to ensure least change on the key-value memories of  several factual prompts. We simply collect the input vectors and output vectors of the selected MLP sublayer with 20,000 prompts from \textsc{COUNTERFACT} which is a subset of the ideal voluminous factual associations dataset that covers as much relations as possible. Notice that we omit the original $C$ in Eqn. 14 for the evaluation of single editing on Llama-2 for all related methods to reduce computational overhead, which has little to no effect on results.

\subsection{G. Evaluation Results on zsRE}
\label{appendix_zsre}
For zsRE, we evaluate the performance on the basic metrics of Efficacy (rewriting accuracy for target editing), Generalization (rewriting accuracy on paraphrased prompts of target editing) and Specificity (maintenance accuracy on unrelated facts). Table 6 displays the evaluation results on zsRE for GPT2-XL. RETS exhibit the competitive performance on zsRE with the baslines, despite that we still deploy the key-value pairs collected from \textsc{COUNTERFACT} for editing. 
\begin{table}[htbp]
\centering
\renewcommand{\arraystretch}{1.2}
\caption{Results of editing zsRE on GPT2-XL.}
\begin{tabular}{cccc}
\hline
Editor & Efficacy & Generalization & Specificity \\ \hline
raw  & 22.9     & 21.9           & 24.2        \\ \hline
FT-L   & 59.52    & \underline{30.9}           & 22.8        \\
MEND   & \underline{20.5}     & \underline{19.9}           & 23.0        \\
ROME   & 99.8     & 87.8           & 24.2        \\
PMET   & 87.3     & 69.6           & 24.2        \\
RETS & 83.5     & 77.9           & 24.2        \\ \hline
\end{tabular}
\label{zsre_re}
\end{table}

\subsection{H. Method Details}
Here we introduce some details for the RETS method, including the optimization process of $v_*^{p_r}$ and the deduction of the solution for $W^{l_e}$. The optimization process of $v_*^{p_r}$ is similar with ROME (Meng et al. 2022a), including maximizing the probability for target object $o^*$ and minimizing the KL divergence of the predictions for the prompt "{subject} is a" to the initial model. The deduction for the solution for $W^{l_e}$ is similar with MEMIT (Meng et al. 2022b).

\subsection{I. Comparison with MEMIT}
For fuller understanding, here we compare RETS with the representative massive editing method but in the case of single editing: MEMIT (Meng et al. 2022b) that spread the update of parameters over several layers. We evaluate on 1,000 samples of the supplemented COUNTERFACT dataset for GPT2-XL. Results are shown in Table 7. It indicates that the subject-focused massive editing method MEMIT also suffers the over-generalizing problem, which shows the deficiency of simply taking subject knowledge into consideration during editing.
\begin{table}[htbp]
\centering
\caption{Performance of single editing for MEMIT and RETS. "raw" stands for the original model before edited. "Gen" stands for Generalization. "S-S" and "R-S" stands for S-Specificity and R-Specificity respectively.}
\renewcommand{\arraystretch}{1.2}
\begin{tabular}{cccccc}
\hline
      & Score & Efficacy & Gen & S-S & R-S \\ \hline
raw   & 55.2  & 20.2     & 22.0           & 78.7          & 100.0         \\
\hline
MEMIT & 78.3  & 94.9     & 79.5           & 77.4          & \underline{61.3}          \\
RETS  & 79.9  & 100.0    & 71.5           & 68.7          & 79.3          \\ \hline
\end{tabular}
\end{table}

\subsection{J. The Trade-off Issue}
We discuss the trade-off issue of RETS compared to the subject-focused methods here. The performance is mainly decided by the editing layer and the editing performance. For the editing layer, the purple line in Figure 5 illustrates the Subject Specificity performance of editing at the last-relation token across layers. It indicates that the subject specificity fluctuates across layers and is even worse in some earlier layers, thus the editing layer is not a major factor for the trade-off of performance. For the editing position, we further analyze the average cosine similarities of $k$ vectors at the last-relation position (input vectors in Equation 13) among subject specificity prompts and generalization prompts respectively. Results are shown in Table 8. $k$ similarity is 0.54 for subject specificity prompts, which is relatively high and indicates little subject information in $k$ vectors; $k$ similarity is 0.39 for generalization prompts. It's not high enough for generalizing to paraphrases, hindering the generalization ability after editing. The above analysis suggests possible disjoint strategies of relation and subject information and joint strategy of paraphrase prompts for the improvement of precise editing.

\begin{table}[ht]
\centering
\caption{The cosine similarities of $k$ vectors of input prompts at the last-relation position. "S-S" stands for S-Specificity and "Gen" stands for Generalization.}
\renewcommand{\arraystretch}{1.2}
\begin{tabular}{ccc}
\hline
             & S-S Prompts & Gen Prompts \\ \hline
$k$ similarity & 0.54                  & 0.39                   \\ \hline
\end{tabular}
\end{table}

\subsection{K. Limitations}
\label{limits}
We notice a few limitations in this work. Firstly, our design of the subject constraints for RETS is relatively simple and affect the quality of text generations. Fluency and Consistency for RETS decline due to the subject constraints that disrupt the associative memories of the original $K$ and $V$ in Eqn. 13, which is verified by the ablation study where Fluency increases sharply without the subject constraints. It calls for future in-depth studies on how subject information is combined into the relation representation. Moreover, our work only applies to editing each factual association separately and it is essential to scale up to sequential or batched facts. Lastly, our work only focuses on the MLP sublayers for knowledge recall following previous works. The further sight into other components may provide a more complete explanation for the inference of the language models.

\end{document}